\makeatletter

\newcommand{\Rmnum}[1]{\expandafter\@slowromancap\romannumeral #1@}
\makeatother

\documentclass[journal]{IEEEtran}
\usepackage{float}
\usepackage{placeins}
\usepackage{siunitx}

\IEEEoverridecommandlockouts

\usepackage{cite}

\ifCLASSINFOpdf
   \usepackage[pdftex]{graphicx}
\else
   \usepackage[dvips]{graphicx}
\fi
\usepackage{xcolor}

\usepackage{multirow}

\usepackage[cmex10]{amsmath}
\usepackage{amsmath,amsfonts,amssymb,amscd,xspace,mathrsfs}
\usepackage{bm}
\usepackage{adjustbox}
\usepackage{setspace}
\usepackage{algorithm}
\usepackage{algpseudocode}
\makeatletter
\def\BState{\State\hskip-\ALG@thistlm}
\makeatother
\usepackage{tikz}
\usetikzlibrary{calc,shapes,3d,decorations,decorations.text, mindmap,shadings,patterns,matrix,arrows,intersections,automata,backgrounds, arrows.meta}

\tikzoption{canvas is xy plane at z}[]{%
	\def\tikz@plane@origin{\pgfpointxyz{0}{0}{#1}}%
	\def\tikz@plane@x{\pgfpointxyz{1}{0}{#1}}%
	\def\tikz@plane@y{\pgfpointxyz{0}{1}{#1}}%
	\tikz@canvas@is@plane
}

\usetikzlibrary{calc}
\usepackage[american,siunitx]{circuitikz}
\usepackage{array}

\ifCLASSOPTIONcompsoc
\usepackage[caption=false,font=normalsize,labelfont=sf,textfont=sf]{subfig}
\else
\usepackage[caption=false,font=footnotesize]{subfig}
\fi

\usepackage{tabularx}

\usepackage{epstopdf}
\usepackage{ifthen}
\usepackage{calc}
\usepackage{makeidx}
\usepackage{algpseudocode}
\algnewcommand{\LeftComment}[1]{\Statex \(\triangleright\) #1}

\newcommand{\meru}[1]{}

\usepackage{footmisc}
\begin{document}
%
\title{Macro-action Multi-time scale \\ Dynamic Programming for Energy Management \\ in Buildings with Phase Change Materials}
%
%
%


\author{\IEEEauthorblockN{Zahra Rahimpour\IEEEauthorrefmark{1},
Gregor Verbi\v{c}\IEEEauthorrefmark{1},
Archie Chapman\IEEEauthorrefmark{2}}\\
\IEEEauthorblockA{\IEEEauthorrefmark{1}%
School of Electrical and Information Engineering,
University of Sydney, NSW, Australia}\\
\IEEEauthorblockA{\IEEEauthorrefmark{2}%
School of Information Technology and Electrical Engineering,
University of Queensland, QLD, Australia}\\
\IEEEauthorblockA{Email:\{zahra.rahimpour, gregor.verbic\}@sydney.edu.au}, archie.chapman@uq.edu.au}

\maketitle

\begin{abstract}
     This paper focuses on energy management in buildings with phase change material (PCM), which is primarily used to improve thermal performance, but can also serve as an energy storage system. In this setting, optimal scheduling of an HVAC system is challenging because of the nonlinear and non-convex characteristics of the PCM, which makes solving the corresponding optimization problem using conventional optimization techniques impractical. Instead, we use dynamic programming (DP) to deal with the nonlinear nature of the PCM. To overcome DP's curse of dimensionality, this paper proposes a novel methodology to reduce the computational burden, while maintaining the quality of the solution. Specifically, the method incorporates approaches from sequential decision making in artificial intelligence, including macro actions and multi-timescale Markov decision processes, coupled with an underlying state-space approximation to reduce the state-space and action-space size. The performance of the method is demonstrated on an energy management problem for a typical residential building located in Sydney, Australia. The results demonstrate that the proposed method performs well with a computational speed-up of up to 12,900 times compared to the direct application of DP.
\end{abstract}



%




\vspace{-0.2cm}
\section{Introduction}
\label{sec:Introduction}

The significant contribution of \textit{heating, ventilation, and air conditioning} (HVAC) in building energy use (up to \SI{50}{\%} in certain countries) and the total contribution of buildings to the overall energy consumption (\SI{20}{\%}-\SI{40}{\%}), make space heating and cooling of growing importance in the area of \textit{demand response} (DR)\footnote{Demand response refers to methods for influencing end-users to use available flexible resources to support network and system services such as load balancing, peak load shaving, and peak load shifting.}  \cite{perez2008review}. 
A potential DR resource available to householders is to use their building's thermal inertia as an energy storage system. However, lightweight buildings, which dominate the residential building stock in Australia and which are the focus of our paper, have low thermal inertia. 
A promising solution to increase their thermal inertia is to use materials with a high heat capacity, such as \textit{phase change materials} (PCM). 
Storing or releasing the latent heat during the phase-change (from solid to liquid or vice versa) provides the building with sufficient thermal mass to smooth indoor temperature fluctuations.

However, to exploit the energy storage capacity of PCM cost-effectively, it needs to be either precooled or preheated (depending on the season) by the HVAC system during shoulder or off-peak hours. This task can be cast as an optimal HVAC scheduling problem, with an objective of minimizing electricity cost while maintaining the indoor temperature  within the desired comfort range.
In the existing literature, this type of optimization problem is classified as a \textit{home energy management} (HEM) problem \cite{keerthisinghe2018fast,Tischer.Verbic.ISGT.2011,DR2}. In spite of the ample literature on the use of PCM for improving thermal performance of buildings \cite{castellon2009experimental,evola2011simulation,konuklu2015review,rahimpour2017using,alam2014energy,Energy,khudhair2008use}, there is a palpable  lack of understanding on how to integrate PCM into HEM, where the non-linear nature of its energy storage can be be exploited using suitable scheduling methods.

To bridge this gap, and in contrast to much of the literature on HEM \cite{keerthisinghe2018fast,Tischer.Verbic.ISGT.2011,DR2}, we consider HEM that consists of an HVAC system as a controllable device and a PCM layer as an energy storage system. 
To date, most HEM optimization problems are solved using \textit{linear programming} (LP) and \textit{mixed integer linear programming} (MILP). 
However, these methods cannot be used to solve nonlinear optimization problems, which phase-change characteristics impart. 
Other methods that are widely used to solve the HEM problems are heuristic methods, such as \textit{particle swarm optimization} (PSO) and \textit{genetic algorithms} (GA). 
The downside of using these methods is that the solution may end up in a local optimum instead of the global optimum, which means the quality of the solution is uncertain \cite{orhean2018new}. 
More importantly, PSO and GA are black-box optimization routines, and in our specific problem they rely on the huge computational task of solving the initial value problems associated with the ordinary differential equations that govern the building's thermal behavior. 
In this sense, they provide no benefit over using principled optimization methods like \textit{dynamic programming} (DP) \cite{bellman1962applied,Tischer.Verbic.ISGT.2011}. 

In this paper, DP is used as the state-of-the-art algorithm for dealing with the nonlinear features of PCM. 
To solve our problem using DP, we first formulate it as a \textit{Markov decision process} (MDP), where the objective is to minimize the accumulated instantaneous cost over a scheduling horizon. 
The main operator in DP is the \textit{value function}, which is formed by summing the expected future costs of following a \textit{policy} (in this problem specific on/off \mbox{sequence} of the HVAC system), given the state transition probabilities. 
Importantly, the objective is equivalent to computing the minimum value function of the problem. 
To do so, the \textit{value iteration} (VI) algorithm is typically employed, which computes the minimum value function in a backward fashion using the Bellman optimality condition\footnote{An optimal policy has the property that whatever the initial state and initial decision are, the remaining decisions must constitute an optimal policy with regard to the state resulting from the first \cite{bellman1962applied}.}. 
However, VI becomes intractable when the time-horizon of the problem, the number of state variables or the number of controllable devices grow. 
In the DP literature, this is known as the \textit{curse of dimensionality} \cite{bellman1962applied}.

As such, in the HVAC-PCM HEM problem, as in many sequential decision problems in \textit{artificial intelligence} (AI), large state-spaces and long time-horizons contribute to a considerable computational challenge. 
In response, the AI literature contains many methods and frameworks for dealing with such large or complex problems. 
Given this, in the next section, we present a brief review of three existing methods from AI that we use to build our computational methodology, namely \textit{state-space approximation}, \textit{multi-timescale Markov decision processes} and \textit{macro actions}. 
It is worth noting that the cornerstone of all these methods is the concept of \textit{abstraction}.
In general, an abstraction is a compact representation of the original problem that is easier to work with than the ground representation. 
Each abstraction method we use works in a different way to reduce the complexity of the HVAC-PCM scheduling problem. 

\subsection{Review of three abstraction approaches in AI}
We now review the three abstraction methods: state-space approximations, multi-timescale MDPs, and macro actions.
   
\textbf{State-space approximations} ---  
The foundation of our method is to discretize the continuous state-space, which is a standard approach to reducing its complexity \cite{rahimpour2019Energy}. 
However, we still left with a large state-space, so we build on this using the following two methods. 
   
\textbf{Multi-timescale Markov decision processes} ---  
The second abstraction approach to reducing the state-space size is to use a multi-timescale MDP, in which decisions are made at different discrete timescales \cite{sutton1995td}. 
Specifically, rather than solving the original MDP as one monolithic problem, we solve several smaller MDPs that are connected successively together to form the original MDP. 
The computation time of the resulting algorithm depends on the choice of each MDP's length, and can be tuned for good performance. 
In our problem, using a multi-timescales model reduces computation time by up to 5,300 times. 
   
\textbf{Macro actions} --- 
The third approach is to use macro actions to reduce the action-space size \cite{pickett2002policyblocks}. 
This approach finds commonalities in the solutions across regions of the state-space to create macro actions, which result in significant computational savings over using the primitive action-space. 
In our methodology, we build on the multi-timescale MDP and treat nearly-identical policies as equivalent. This itself reduces the number of policies to consider by one third and speeds up the performance of the algorithm further by 2.4 times. 
However, the major limitation of applying macro actions is the quality of the solution. 
Certain policies cannot be captured since, in the macro-action MDP, the policies contain only macro actions.
Therefore, the resulting policy may be suboptimal \cite{hauskrecht1998hierarchical}; however, this results in poor performance only if the macro actions themselves are poorly selected \cite{precup1998theoretical}.  
   
\subsection{Contributions of the paper}
The main technical contribution of this paper is to develop a computationally efficient algorithm for scheduling a controllable HVAC system in buildings with PCM. 
To our knowledge, this work is the first attempt to solve an optimization problem in buildings with PCM.
Beyond this, the paper advances the state of the art in the following ways:
   \begin{enumerate}
   	\item We derive a novel and computationally-efficient optimization method for online non-linear scheduling, which exploits several techniques from AI in one framework. 
   	\item We demonstrate the method on a HVAC scheduling problem in a typical PCM-building, incorporating a non-linear RC lumped thermal model.
   	\item We evaluate the method on four different seasonal weather conditions, with results showing that the method has good accuracy over DP, and provides a considerable electricity cost saving over a deadband controller.
   	\item The proposed method gives substantial computational speed-ups, as much as 12,900 times faster than DP.
   	\item The method described in this paper can be implemented on current smart meters and IoT gateway devices, such as those built on Raspberry Pi boards.
   \end{enumerate}
   
\subsection{Outline of the paper} 
   
This paper progresses as follows: in Section \ref{sec:Thermal RC lumped model of PCM-buildings}, a thermal model of a PCM-building is built in MATLAB, and validated by benchmarking it against an identical model in EnergyPlus\footnote{EnergyPlus is a software tool that is widely used for simulating the thermal behavior of buildings.}. 
In Section \ref{sec:Home energy management in PCM buildings}, the optimization problem of the PCM-building is formulated as an MDP, and the \textit{value iteration} algorithm is used to solve it. 
Section~\ref{sec:Methodology} contains the main technical contributions of this paper, where the proposed methodology of macro-action multi-timescale dynamic programming is derived.
In Section~\ref{sec:Implementation}, the method is implemented on a typical PCM-building in Sydney, and its performance is evaluated over four seasonal weather conditions. 
Section~\ref{sec:Conclusion} concludes and outlines future directions of this work.

\section{Thermal RC lumped model of PCM-buildings}       
\label{sec:Thermal RC lumped model of PCM-buildings}       
\par To formulate the HVAC-PCM optimization problem, we need a thermal model of the building that strikes the right balance between accuracy and computational efficiency. Therefore, we first build a thermal model of a PCM-building in MATLAB, which we then evaluate by benchmarking it against an identical model in the EnergyPlus software.

\subsection{Thermal model of PCM-building } 

\begin{figure}[t]
	\centering
	\begin{circuitikz} [american voltages]
		\draw (0,0) to [short] (0,1.0) to [R=$R_\text{dw}$] (4,1.0) to [short] (4,0);
		\draw (4,0) to[C=$m_\text{a}$$c_\text{a}$,-] (4,-1.0) node[ground]{};
		\draw (0,0) to[R=$R_\text{out}$,o-o] (2,0) to [R=$R_\text{in}$,o-o] (4,0);
		\draw[{Latex[length=3mm, width=2mm]}-] (4.1,0.2) -- (4.8,1.1) node[right] {$Q_\text{HVAC}+Q_\text{inf}$};
		
		\draw (2,-1.0) node[ground]{} to[C=$C_\text{e}$+$C_\text{PCM}$,-] (2,0)
		(0,0) node [left]{$T_\text{out}$}
        (2,0) node [above]{$T_\text{e}$}
		(4,0) node [right]{$T_\text{in}$};
		
	\end{circuitikz}
	\caption{Thermal lumped parameter model of a PCM-building.}
	\vspace{-0.2cm}  
	\label{2RC}
\end{figure}
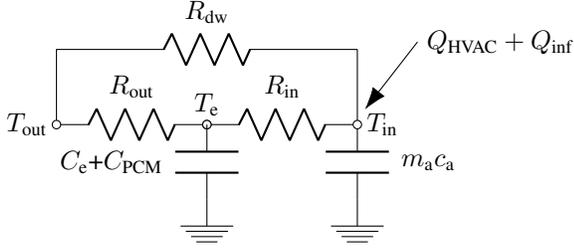

A simple way of modeling the thermal performance of a building is using an RC  representation of each building element, which is known as an RC lumped model\cite{underwood2008modelling,gouda2000low}. 
For simplicity, we use a 2RC model, where all elements of the wall, roof, and floor are lumped together into two lumped resistances and one lumped capacitance, as shown in Fig.~\ref{2RC}. The circuit is thermally excited by the internal heat flow, including the HVAC system $Q_\mathrm{HVAC}$, and natural air infiltration $Q_\mathrm{inf}$, and the outside temperature ${T_\mathrm{out}}$. The building’s interior is simulated as a single capacitor $C_\mathrm{a}$, which captures the heat capacity of air. Elements like doors and windows have negligible thermal inertia, so they are modeled using resistance $R_\mathrm{dw}$ in parallel with the lumped model of the wall. 

The dominant capacitance in the model is $C_\mathrm{e}$+$C_\mathrm{PCM}$, which captures the combined the thermal inertia of the envelope and the PCM layer.
The building's envelope is made up of three layers: rendered fibro-cement, a timber stud wall containing insulation batts, and plasterboard on the inside\footnote{The materials and configuration are chosen based on a  common practice in lightweight building in Australia \cite{gregory2008effect}.}. To improve the thermal inertia, PCM is added as a \SI{0.03}{\metre} layer underneath the plasterboard. Therefore, two layers with significant amount of thermal mass (PCM layer and timber wall) are placed adjacent and can be lumped together as a single capacitance.
The parameters are chosen so that the model represents a single-zone $\SI{8}{\metre}\times\SI{6}{\metre}\times\SI{2.7}{\metre}$ cuboid with a total floor area of $\SI{48}{\metre\squared}$. The detailed calculations of the resistances and capacitances of the model can be found in \cite{rahimpour2018APVI}. 

\begin{figure}[t]
	\centering
	\includegraphics[width=70mm,keepaspectratio]{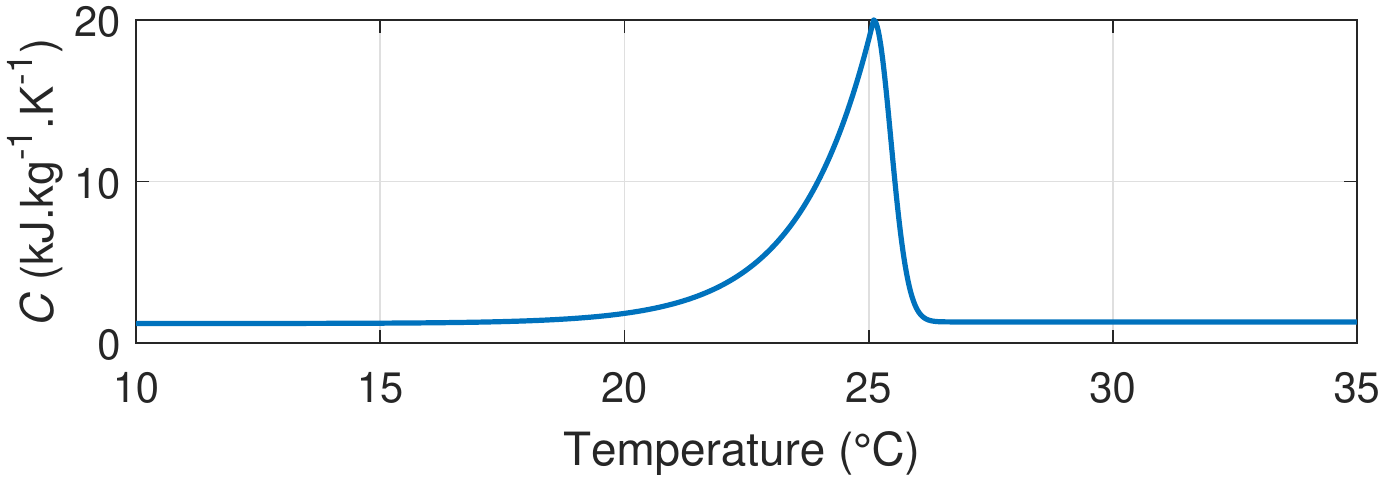}
	\caption{Specific heat capacity characteristic of a typical PCM.}
	\label{cpcm}
\end{figure}

The specific heat capacity characteristic of the PCM used in this paper is shown Fig.~\ref{cpcm}. Observe that the phase change occurs over the range between \SI{20}{\degreeCelsius} and \SI{26}{\degreeCelsius}. The melting point is at \SI{25.1}{\degreeCelsius}, where the specific heat capacity is the largest. The PCM melting point is an important design parameter, chosen to reflect the occupants' comfortable temperature range.
The PCM specific heat capacity is given by:
\begin{subequations} \label{eqn1}
	\begin{align}
	\begin{split} \label{eqn1a}
	{c_\mathrm{pcm}}&= 1200 + 18800e^{-\left(\frac{{T_\mathrm{p}} - {T}}{1.5}\right)}  \quad \mathrm{if} \quad  {T} < {T_\mathrm{p}},   \\
	\end{split}\\
	\begin{split} \label{eqn1b}
	{c_\mathrm{pcm}}&= 1300 + 18700e^{-4\left({T_\mathrm{p}} - {T}\right)^{2}}  \quad \mathrm{if} \quad  {T} \geq {T_\mathrm{p}},  \\
	\end{split}
	\end{align}
	\label{straincomponent}
\end{subequations}
\hspace{-1.2em} where ${T_\mathrm{p}}$ is the melting point of the PCM. 
Finally, the thermal model of a PCM building consists of two first-order differential equations capturing the energy balance at the two nodes defined by, respectively, the indoor temperature ${T_\mathrm{in}}$ and the surface temperature of the PCM layer ${T_\mathrm{e}}$:
\begin{equation}\label{eqn7}
{\dot{T_\mathrm{e}} = \frac{1}{C_\mathrm{e}+C_\mathrm{PCM}}\left(\frac{T_\mathrm{in} - T_\mathrm{e}}{R_\mathrm{in}}+\frac{T_\mathrm{out} - T_\mathrm{e}}{R_\mathrm{out}}\right)},
\end{equation}
\begin{equation}\label{eqn8}
{\dot{T}_\mathrm{in} = \frac{1}{m_\mathrm{a}c_\mathrm{a}}\left(\frac{T_\mathrm{out}\!-\! T_\mathrm{in}}{R_\mathrm{dw}}+\frac{T_\mathrm{e}\!-\! T_\mathrm{in}}{R_\mathrm{in}}\!+\!\dot{Q}_\mathrm{HVAC}\!+\!\dot{Q}_\mathrm{inf}\right)}.
\end{equation}
Note that due to the discontinuity in \eqref{eqn1a} and \eqref{eqn1b}, the PCM heat capacity curve needs to be fitted with a polynomial function before it can be used to numerically solve \eqref{eqn7}.

\subsection{Benchmarking thermal model against EnergyPlus}
\begin{figure}[t]
	\centering
    \includegraphics[width=88mm,keepaspectratio]{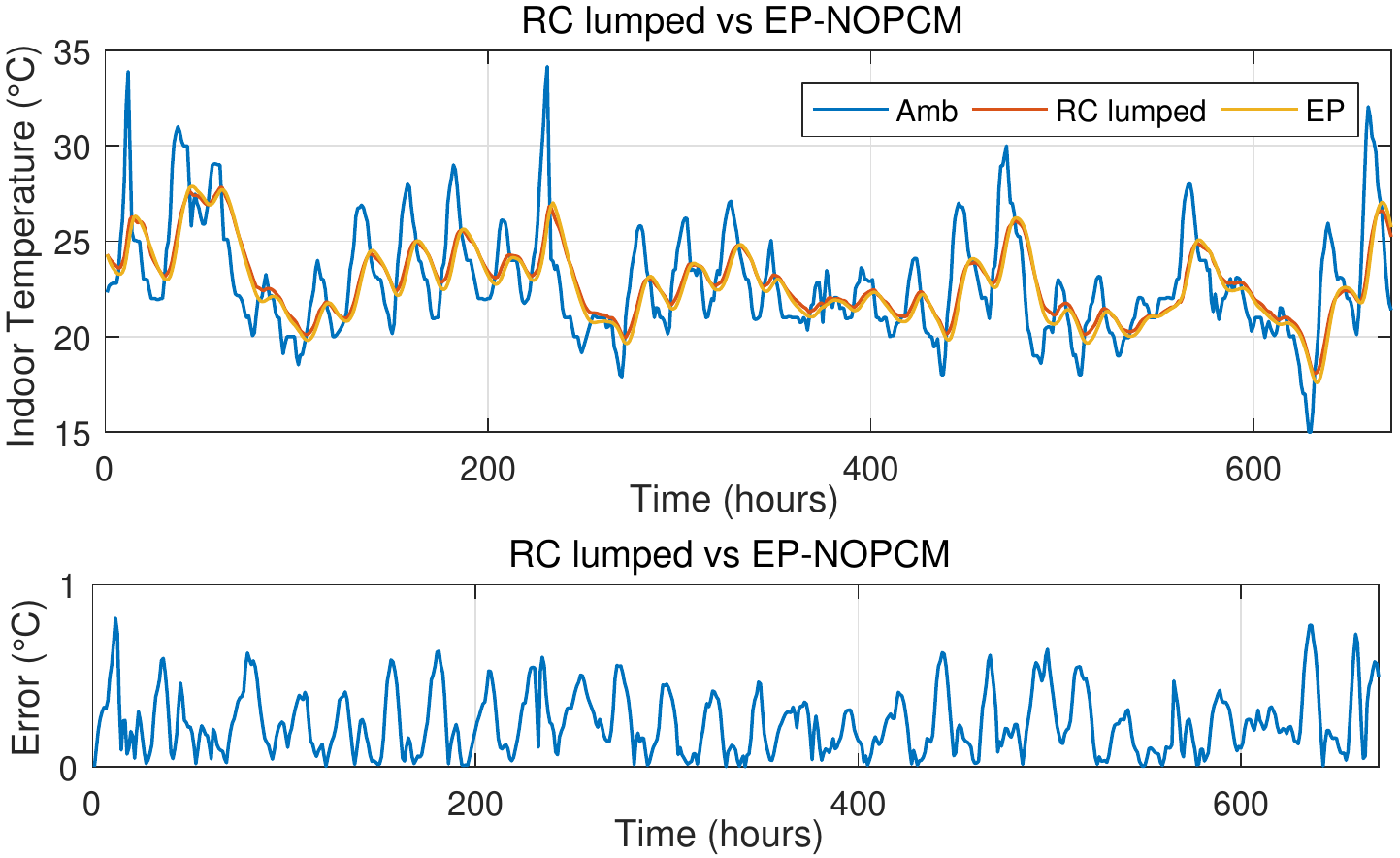}
	\caption{Indoor temperature of a building without PCM using RC lumped model (RC lumped) benchmarked against EnergyPlus results (EP) and compared to the ambient temperature (Amb).}
	\vspace{-0.2cm} 
	\label{figa}
\end{figure}

\begin{figure}[t]
	\centering
	\includegraphics[width=88mm,keepaspectratio]{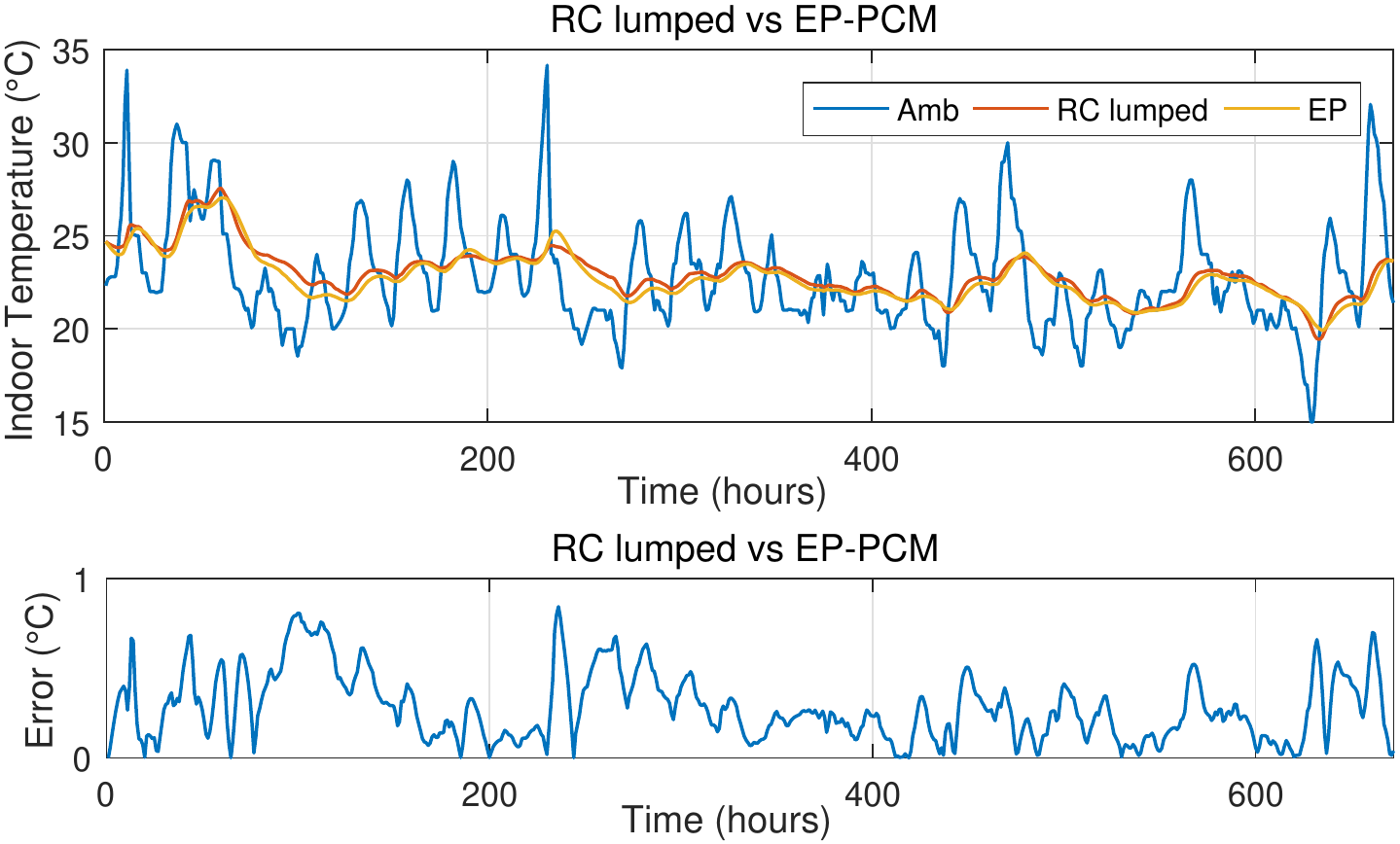}
	\caption{Indoor temperature of a PCM-building using RC lumped model (RC lumped) benchmarked against EnergyPlus results (EP) and compared to the ambient temperature (Amb).}
	\vspace{-0.2cm}
	\label{figb}
\end{figure}

We now compare the performance of the RC lumped model implemented in MATLAB against an identical model in EnegyPlus, using  the indoor temperature of the building for comparison. 
Simulations are performed for a typical summer month (1--28 February) in Sydney. We use the root-mean-square error (RMSE) as comparison metric. 
Observe in Figs.~\ref{figa} and \ref{figb} that the models match well in both scenarios, with and without PCM, with the maximum RMSE value not exceeding \SI{0.8}{\degreeCelsius}; this is acceptable with respect to the model uncertainty and human temperature sensitivity.

\section{Home energy management in PCM buildings} 
\label{sec:Home energy management in PCM buildings}
In this section, we present an MDP formulation of the HVAC-PCM optimization problem using differential equations \eqref{eqn7} and \eqref{eqn8} of the validated model as transition functions. Then we show how to use DP to solve the optimization problem.

An MDP comprises a \textit{state-space}, $(s\in \mathcal{S})$, a \textit{decision-space}, $(x\in \mathcal{X})$,  \textit{transition functions} and \textit{contribution functions}. 
Let $k=\left\lbrace 1,\ldots,K \right\rbrace$ denote a time-step of one hour. A state variable, $s_{k}\in \mathcal{S}$, contains the information that is necessary and sufficient to make the decisions and compute costs, rewards and transitions. 
The decision variable, $x_{k}\in \mathcal{X}$, is an action that results in a transition from one state to another, in a sequence over the decision horizon. 
Finally, random effects are in general used to represent chance exogenous information, such as weather conditions or inhabitants' behavioural patterns \cite{keerthisinghe2018fast}. However, for simplicity and because we focus on the non-linear characteristics of PCM, in this work the problem is treated as deterministic, and as such, random effects are omitted and left for future work. 
Thus, the form of the HVAC-PCM MDP is given by
\begin{align}\label{eqn22}
\underset{\pi}{\text{min}}\ 
& \mathbb{E}\left\{ \sum^{K}_{k=0} C_{k}({s}_{k},{x}_{k} = \pi(s_k))\right\} \nonumber \\
\text{s.t.} 
&\quad \textrm{thermal comfort constraints, and} \nonumber \\ 
&\quad \textrm{thermal energy balance constraints,}
\end{align}
where $\pi:\mathcal{S}\rightarrow \mathcal{X}$ is a policy, i.e. a sequence of actions taken to move from each state to the next state over the whole time horizon. In this work, a policy is a sequence of on/off status of the HVAC system over a defined time horizon. 

The function 
$C_{k}({s}_{k},{ x}_{k})$
is the contribution function, which is the cost incurred at a given time-step $k$ that accumulates over time \cite{keerthisinghe2018fast}.
For our specific optimization problem, the cost consists of the electricity cost and the discomfort cost:
\begin{equation}\label{eqn23}
C_{k}({s}_{k},{x}_{k})
=
\lambda{c}_{\mathrm{g},k}{P}_{k} + \left(1-\lambda\right) \left(\mathopen|T_{\mathrm{room},k}- T_\mathrm{s}\mathclose|\right).
\end{equation}

To balance the two cost components, the contribution function includes a weighting factor ${\lambda}$, applied to the electricity cost, with ($1-{\lambda}$) applied to the penalty for deviating from the desired HVAC set point ${T_\mathrm{s}}$. 
We assume a reverse-cycle HVAC system able to operate both in a heating and cooling mode. The setpoint ${T_\mathrm{s}}$ for the two modes is assumed \SI{20}{\degreeCelsius} and \SI{23}{\degreeCelsius}, respectively.
The electricity cost of the HVAC system is the electricity time-of-use tariff, (${c}_{\mathrm{g},k}$), multiplied by the energy used to run the HVAC system, ${P}_{k}$.

Referring back to \eqref{eqn22}, let ${{s}_{k+1}={s}^{M}\left({s}_{k},{x}_{k}\right)}$ describe the state evolution from time step $k$ to the next time step, $k+1$, where ${s}^{M}$ is the underlying mathematical model of the studied system \cite{keerthisinghe2018fast}. In this problem, the system model is the thermal model of the building, so the MDP transition functions are given by \eqref{eqn7} and \eqref{eqn8}. 

Cost function \eqref{eqn23}, only considers the instantaneous cost that results from the decision that is taken at each time step. 
Building on this, DP solves the optimization problem by computing a \textit{value function} $V^{\pi}({s}_{k})$, which is the expected future discounted cost of following policy $\pi$ starting in state ${s}_{k}$. It is given by:
\begin{align}\label{eqn24}
{\displaystyle V^{\pi}({s}_{k})
= \sum_{{s}'\in \mathcal{S}} \mathrm{Pr}({s}'|{s}_{k},{x}_{k})\left[C({s}_{k},{x}_{k},{s}')+ V^{\pi}({s}')\right]}, 
\end{align} 
where $\mathrm{Pr}({s}'|{s}_{k},{x}_{k})$ is the transition probability of landing on state ${s}'$ from ${s}_{k}$ if we take action $x_k$ \cite{keerthisinghe2018fast}. 
However, because the system model, $s^M$ is a deterministic function, we have
\begin{equation}\label{eqn240}
\mathrm{Pr}({s}'|{s}_{k},{x}_{k}) = 
\begin{cases}
  1 & \textrm{if } s^M(s_k,x_k) = s', \\
  0 & \textrm{otherwise,}
\end{cases}
\end{equation}
making the calculation simpler than in the stochastic case.

The expression in \eqref{eqn24} is a recursive reformulation of the objective function. 
Thus, in general, \textit{Bellman's optimality condition} states that the optimal value function is given by
\begin{align}\label{eqn25} 
V_{k}^{\pi^{*}}({s}_{k})=\min\limits_{{x}_{k}\in \mathcal{X}_{k}} \left(C_{k}({s}_{k}, {x}_{k}) + \mathbb{E}\left\{ V_{k+1}^{\pi^{*}}({s}')|{s}_{k}\right\}\right),
\end{align}
where $\pi^{*}$ is an optimal policy.
To find $\pi^*$, we need to solve \eqref{eqn25} for each state. 

\textit{Value iteration} (VI) is the process of computing \eqref{eqn25} for each state by backward induction; that is, starting at the end points of the MDP. 
The optimal policy is extracted from the optimal value function by selecting the minimum value action for each state. 
To describe this in a simple way, in VI, the desired state in step $k+1$ is set to the lower value while the undesired states and states that are out of comfort bounds are penalized by assigning higher values. 
Then, for all possible states at time $k$, the VI algorithm moves backward in time and, in each time step, by solving the subproblem in~\eqref{eqn25}, the minimum value function is computed for different states of each time step. 
In the final step of backward induction, corresponding to the initial starting point, all value function calculations converge to the optimal value function. 
Then, by tracing a minimum value-function path forward for a given time horizon, the optimal policy is found \cite{keerthisinghe2018fast}. 

However, despite advancements in computation power, directly applying VI (or other exact DP algorithms) has an excessively high computational burden.  
Although we consider only one state-variable representing the indoor temperature of the building, the running time of the VI algorithm for a decision horizon of 24 hours with slot length of one hour is very long; in this problem, the running time is almost nine days on a high-performance computer cluster. 
The main reason behind this is that at each time step, the algorithm solves differential equations \eqref{eqn7} and \eqref{eqn8} for each action to update the MDP state, and this is repeated until the initial starting state is reached. 
Given this shortcoming, we now propose a method to overcome the computational burden of the DP algorithm. 

\section{Methodology}       
\label{sec:Methodology}
In this section, we describe our methodology in three steps, namely: 
state-space approximation, multi-time scale MDP, and macro-action abstraction. 
 
\subsection{State-space approximation}
We now describe the state-space approximation used as a first step to deal with the computational burden of DP.
However, before explaining the methodology, a few terms need to be defined. 
We call the MDP that uses equations \eqref{eqn7} and \eqref{eqn8} without any change, the \textit{exact model}. 
The output of the exact model in each time-step is a state (indoor temperature), which we call the \textit{exact state}. If we consider all the possible states over the decision horizon, we call that the \textit{exact state-space}. 
The corresponding terms in the approximated methodology are called, respectively, the \textit{approximate model}, \textit{approximated state} and \textit{approximated state-space}. 

The proposed approximation involves rounding the output of differential equations~\eqref{eqn7} and~\eqref{eqn8} in each time step to the nearest multiple of 0.1. 
Given this, depending on the state trajectory of the exact model, an approximate state may cover more than one exact state, which improves the computational performance of the VI algorithm by reusing the computed state transitions and sub-problem solutions. 
In more detail, for the desired comfort range between \SI{20}{\degreeCelsius} and \SI{26}{\degreeCelsius} and assuming a penalty for out-of-bound temperatures, any state in the state-space is a value between \SI{20}{\degreeCelsius}  and \SI{26}{\degreeCelsius} with a \SI{0.1}{\degreeCelsius} discretization. 
In other words, we can group the whole desired state-space into 61 groups. This approximation is acceptable as long as it does not affect the quality of the optimization solution. This is demonstrated in the first part of Section~\ref{sec:Implementation}, by defining metrics to measure the quality of the solution resulting from this state-space approximation. 
The approximated MDPs that are developed in this section, will be used as a ground MDP for the multi-time scale MDP and macro-action abstraction. 


\vspace{-0.2cm}  
\subsection{Multi-timescale Markov decision processes}  
\label{subsecc:Multi-timescale Markov decision processes}
Applying a multi-timescale abstraction significantly reduces the computational burden of energy management optimization problem. 
Building on the approximated state-space introduced above, we divide the time-horizon of the problem into \textit{blocks}, each consisting of four time-steps (hours). 
We note that the performance of our methodology highly depends on the length of each block.
Through the process of trial and error, we found that choosing four time-steps for each block strikes the right balance between the number of blocks and the length of each block, and results in the highest speed-up. 
For reference, we denote the  multi-timescale method Algorithm~1 (\textsc{Alg~1}). 
We can formulate each block as a separate MDP; therefore, we solve a few successive block-MDPs using VI to find the optimal policy over the whole time-horizon.  

\begin{algorithm}[!t]
\caption{: \small Multi-time scale algorithm (\textsc{Alg~1})}\label{alg1}
\begin{algorithmic}[1]
\Statex $L$: length of each block
\Statex ${T_\mathrm{1}}$: lower bound of desired temperature range
\Statex ${T_\mathrm{2}}$: upper bound of desired temperature range
\Statex $d$: discretization step
\Statex ${T_\mathrm{0}}$: fix initial temperature
\Statex \small \LeftComment Value iteration (VI) of the last block ($\mathrm{Blk}_{M}$)
\begin{spacing}{0.8}
\end{spacing}
\For  {\small ${T_\mathrm{1}},...,{T_\mathrm{2}}$ \quad{with $\mathrm{d}$ step discretization}} 
\For{ all ${2^\mathrm{L}}$ combinations of the action-space}
\State \small calculate states using equations \eqref{eqn7} and \eqref{eqn8}
\EndFor\label{timeendwhile}
\If {${T_\mathrm{1}}\leq {s}_\mathrm{Blk_{M,colL}}\leq {T_\mathrm{2}}$} 
\State \small Initialize $v_{Blk_{M,colL}}$ to zero vector
\Else
\State \small Initialize $v_\mathrm{Blk_{M,colL}}$ to infinity vector.
\EndIf
\State \small Execute the VI and store final value function in $v_\mathrm{final,Blk_{M}}$.
\EndFor
\Statex \small \LeftComment Value iteration of Blocks 2 to $M-1$ ($\mathrm{Blk}_{2}$ to $\mathrm{Blk}_{M-1}$).
\begin{spacing}{0.8}
\end{spacing}
\For{$M-1:...:2$}
\For{${T_\mathrm{1}},...,{T_\mathrm{2}}$ \quad{with $\mathrm{d}$ step discretization}}
\For{ all ${2^\mathrm{L}}$ combinations of the action-space}
\State \small calculate states using equations \eqref{eqn7} and \eqref{eqn8}
\EndFor\label{timeendwhile}
\If {${T_\mathrm{1}}\leq {s}_\mathrm{Blk_{M-1,colL}}\leq {T_\mathrm{2}}$}
\State \small for any state for which {$s_\mathrm{Blk_{M-1,colL}}=s_\mathrm{Blk_{M,col1}}$},
\State \small Initialize $v_\mathrm{Blk_{M-1,colL}}$ with the corresponding $v_\mathrm{final,Blk_{M}}$. 
\Else 
\State \small Initialize $v_\mathrm{Blk_{M-1,colL}}$ with infinity vector
\EndIf
\State \small Execute the VI and store final value function in ${v}_\mathrm{final,Blk_{M-1}}$.
\EndFor
\EndFor  
\Statex \small \LeftComment Value iteration of the first block ($\mathrm{Blk}_{1}$).
\begin{spacing}{0.8}
\end{spacing}
\State \small Set ${T_\mathrm{0}}$. 
\For{ all ${2^\mathrm{L}}$ combinations of the action-space}
\State \small calculate states using equations \eqref{eqn7} and \eqref{eqn8}
\EndFor\label{timeendwhile}
\If {${T_\mathrm{1}}\leq {s}_\mathrm{Blk_{1,colL}}\leq {T_\mathrm{2}}$}
\State \small for any state for which {$s_\mathrm{Blk_{1,colL}}=s_\mathrm{Blk_{2,col1}}$},
\State \small Initialize $v_\mathrm{Blk_{1,colL}}$ with the corresponding $v_\mathrm{final,Blk_{2}}$. 
\Else 
\State \small Initialize $v_\mathrm{Blk_{1,colL}}$ with infinity vector
\EndIf
\State \small Execute the VI and store final value function in ${v}_\mathrm{final,total}$.
\end{algorithmic}
\end{algorithm}

As described in the pseudocode of \textsc{Alg}~1, first we apply VI algorithm on the last block-MDP (Lines 1-11). In more detail, we set the corresponding value functions in the last time step to zero for the states that have a value within the desired comfort range, and assign a high value for the states with the values out of the comfort range (Lines 5-9).
To exploit the advantage of the approximated state-space, we run the VI algorithm for 61 (20:0.1:26) initial points (Lines 1 and 13) for each block-MDP except the first one. We save all the optimal value functions that correspond to each of the 61 initial points in look-up tables (Lines 10 and 23). Before running VI on the remaining block-MDPs, we update the initial value functions by replacing the corresponding value function of the current state. In more detail, we find the initial states that have the same value as the current state and replace the corresponding value functions as the initial value functions of the current block-MDP (Lines 18-19 and Lines 31-32).  
 This process repeats until the first block-MDP. 
To find a solution for the optimization problem over a defined time-horizon, we need to fix either the initial or the final temperature.
In this work, we provide the \textsc{Alg}~1 with a fixed initial temperature (Line 26). 
Therefore, we have only one VI to run for the first MDP. 

Comparing the results of \textsc{Alg}~1 with a one-block MDP model shows that both methods converge to the exactly same solution. 
This corroborates with Sutton's result that $n$-block MDPs act exactly same as the corresponding one-block MDP \cite{sutton1995td}. 
{The mathematical proof of this claim, tailored to our setting, is given in the online Appendix.}
Our simulations show that using \textsc{Alg}~1 reduces the computational burden of finding the optimal policy by a factor of 5,300.

\tikzset{every picture/.style={line width=0.75pt}} 
\begin{figure}
	\centering
	
\tikzset{every picture/.style={line width=0.75pt}} 

\begin{tikzpicture}[x=0.75pt,y=0.75pt,yscale=-1,xscale=1,scale=0.90,every node/.style={scale=0.90} ]

\draw   (70,150) -- (70,100) -- (420,100) -- (420,150) -- cycle ;
\draw  [dash pattern={on 0.84pt off 2.51pt}]  (90,150) -- (90,100) ;

\draw  [dash pattern={on 0.84pt off 2.51pt}]  (180,150) -- (180,100) ;

\draw  [dash pattern={on 0.84pt off 2.51pt}]  (310,150) -- (310,100) ;

\draw  [dash pattern={on 0.84pt off 2.51pt}]  (400,150) -- (400,100) ;

\draw   (90.43,151.47) .. controls (90.43,156.14) and (92.76,158.47) .. (97.43,158.47) -- (124.43,158.47) .. controls (131.1,158.47) and (134.43,160.8) .. (134.43,165.47) .. controls (134.43,160.8) and (137.76,158.47) .. (144.43,158.47)(141.43,158.47) -- (171.43,158.47) .. controls (176.1,158.47) and (178.43,156.14) .. (178.43,151.47) ;
\draw   (310.43,97.47) .. controls (310.43,92.8) and (308.1,90.47) .. (303.43,90.47) -- (255.43,90.47) .. controls (248.76,90.47) and (245.43,88.14) .. (245.43,83.47) .. controls (245.43,88.14) and (242.1,90.47) .. (235.43,90.47)(238.43,90.47) -- (187.43,90.47) .. controls (182.76,90.47) and (180.43,92.8) .. (180.43,97.47) ;
\draw   (311.43,152.47) .. controls (311.47,157.14) and (313.82,159.45) .. (318.49,159.41) -- (345.99,159.18) .. controls (352.66,159.12) and (356.01,161.42) .. (356.05,166.09) .. controls (356.01,161.42) and (359.32,159.06) .. (365.99,159.01)(362.99,159.03) -- (393.49,158.78) .. controls (398.16,158.74) and (400.47,156.39) .. (400.43,151.72) ;
\draw   (89.43,96.47) .. controls (89.43,94) and (88.2,92.76) .. (85.73,92.76) -- (85.73,92.76) .. controls (82.2,92.76) and (80.43,91.52) .. (80.43,89.05) .. controls (80.43,91.52) and (78.67,92.76) .. (75.14,92.76)(76.73,92.76) -- (75.14,92.76) .. controls (72.67,92.76) and (71.43,94) .. (71.43,96.47) ;
\draw   (419.93,94.42) .. controls (419.93,91.81) and (418.63,90.5) .. (416.02,90.5) -- (416.02,90.5) .. controls (412.29,90.5) and (410.43,89.2) .. (410.43,86.59) .. controls (410.43,89.2) and (408.57,90.5) .. (404.85,90.5)(406.52,90.5) -- (404.85,90.5) .. controls (402.24,90.5) and (400.93,91.81) .. (400.93,94.42) ;

\draw (86.5,77) node [scale=0.7]  {$\phi =0$};
\draw (134,173) node [scale=0.7]  {$\phi =0.25$};
\draw (241,77) node [scale=0.7]  {$\phi =0.5$};
\draw (356,173) node [scale=0.7]  {$\phi =0.75$};
\draw (406.5,77) node [scale=0.7]  {$\phi =1$};
\draw (126.59,124.98) node [scale=0.7  ,opacity=1 ,rotate=-0.21] [align=left] {0\\0\\1 \\0};
\draw (106.5,125) node [scale=0.7  ,opacity=1 ] [align=left] {0\\0\\0 \\1};
\draw (196.5,125) node [scale=0.7  ,opacity=1 ] [align=left] {0\\0\\1 \\1};
\draw (146.5,125) node [scale=0.7  ,opacity=1 ] [align=left] {0\\1\\0 \\0};
\draw (216.5,125) node [scale=0.7  ,opacity=1 ] [align=left] {0\\1\\0 \\1};
\draw (236.5,125) node [scale=0.7  ,opacity=1 ] [align=left] {0\\1\\1\\0};
\draw (326.5,125) node [scale=0.7  ,opacity=1 ] [align=left] {0\\1\\1\\1};
\draw (166.5,125) node [scale=0.7  ,opacity=1 ] [align=left] {1\\0\\0 \\0};
\draw (256.5,125) node [scale=0.7  ,opacity=1 ] [align=left] {1\\0\\0\\1};
\draw (276.5,125) node [scale=0.7  ,opacity=1 ] [align=left] {1\\0\\1\\0};
\draw (346.5,125) node [scale=0.7  ,opacity=1 ] [align=left] {1\\0\\1\\1};
\draw (296.5,125) node [scale=0.7  ,opacity=1 ] [align=left] {1\\1\\0\\0};
\draw (366.5,125) node [scale=0.7  ,opacity=1 ] [align=left] {1\\1\\0\\1};
\draw (413.5,125) node [scale=0.7  ,opacity=1 ] [align=left] {1\\1\\1\\1};
\draw (386.5,125) node [scale=0.7  ,opacity=1 ] [align=left] {1\\1\\1\\0};
\draw (83.5,125) node [scale=0.7  ,opacity=1 ] [align=left] {0\\0\\0 \\0};
\end{tikzpicture}
\caption{The combinatorial structure of each optimal policy.}
\vspace{-0.2cm}
\label{fig:M1}
\end{figure}
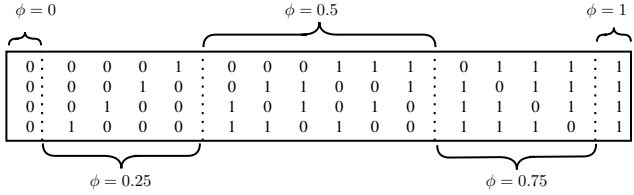

\subsection{Macro actions}  

Building on \textsc{Alg}~1, we now describe the macro-action abstraction. 
This is a widely-used method for reducing the size of a state-space. 
After running VI on each block-MDP for different initial points and observing the corresponding state-spaces, we define the macro actions. 
Specifically, the contribution function of the ground MDP \eqref{eqn23} is modified to:
\begin{equation}\label{eqn33}
\bar{C}^{\pi}_{k}(\bar{s}_{k},\bar{x}_{k})
=
\lambda\: {c}_{\mathrm{g},k}{P}_{k}{\phi} + \left(1-\lambda\right) \left(\mathopen|T_{\mathrm{room},k}- T_\mathrm{s}\mathclose|\right),
\end{equation}
where $\phi:\mathcal{S}\rightarrow \mathcal{\bar{S}}, {\phi} \in \{\,0,0.25,0.5,0.75,1\,\}$ is a percentage of the HVAC system rated power used for the abstraction, and $\mathcal{\bar{S}}$ is the abstract state-space. 
To be clear, each ground MDP over four-time steps has ${2}^{4}$ possible policies consisting of primitive actions. 
These 16 policies are all the possible on/off permutations of the HVAC system over four-time steps. 
Under our defined macro actions, all 16 combinatorial arrangement of zero, one, two, three and four hours of operation of the HVAC system have the corresponding compact representations of no operation, or four hours operation of HVAC system at \SI{25}{\%}, \SI{50}{\%}, \SI{75}{\%} and \SI{100}{\%} rated power, respectively.  
This is illustrated in Fig.~\ref{fig:M1}. 

To this end, we implement our methodology in two phases, and call it Algorithm~2 (\textsc{Alg}~2).  
In the first phase, we implement \textsc{Alg}~1 on MDPs with macro actions. 
These solutions differ from those of \textsc{Alg}~1, which is applied to the MDP with primitive actions (Section \ref{subsecc:Multi-timescale Markov decision processes}). 
Then, in the second phase, we expand the optimal macro-policies for the next-executed block-MDP to their related primitive policies, as shown in Fig.~\ref{fig:M1}.  
Sequentially expanding the optimal policy of these new MDPs gives the overall policy computed by \textsc{Alg}~2.
Our simulation results in Section~\ref{sec:Implementation} show that policies produced by \textsc{Alg}~2 are close to those found with \textsc{Alg}~1, so they can be assumed near-optimal. 

\section{Evaluation and discussion} 
\label{sec:Implementation}
We begin by evaluating the loss in the solution quality resulting from the state-space approximation.
We then examine the quality of the policies computed using, respectively, \textsc{Alg}~1 and \textsc{Alg}~2, and compare their computational performance.

We define three measures of the solution quality, which we use in Section~\ref{subsec:State-space approximation}. 
These are: 
(i) \textit{mean-absolute error} (MAE) of the approximated state-space against the exact state-space (i.e. the temperature error), 
(ii) MAE of the final value function resulting from applying VI on the approximated state-space versus the final value function resulting from applying VI on the exact state-space without temperature discretization, and
(iii) normalized \textit{calibration error} between the optimal policy using DP on the approximated state-space compared to the optimal policy using DP on the exact state-space\footnote{Here we use the term \textit{calibration} in the statistical sense, to measure the fit of the approximate DP method to exact DP.}.
In Section~\ref{subsec:Alg1vsAlg2}, similar measures are used to compare the results of \textsc{Alg}~1 and \textsc{Alg}~2. 

The simulations of the Section~\ref{subsec:Alg1vsAlg2} were run in MATLAB using a computing platform with an Intel 2.7 GHz i7-7500U CPU, 64-bit operating system and 16 GB RAM, while for Section V-A, we used a high-performance computer cluster due to the excessive computational burden.  

\subsection{Benefits of the state-space approximation}
\label{subsec:State-space approximation}
As mentioned in the previous section, the runtime of the exact VI for a time horizon of 24 hours is nine days. 
Thus, we consider the benefits of using state-space approximation, and calculate the performance metrics, for only one typical summer day in Sydney.
To have a reasonable population to assess the approximation by the three criteria above, we generated both the approximated and the exact state-space for the 61 initial points between \SI{20}{\degreeCelsius} and \SI{26}{\degreeCelsius} with a \SI{0.1}{\degreeCelsius} discretization.

First, we calculate the MAE of the approximated state-space, which represents the indoor temperature versus the actual state-space over a time horizon of 24 hours. The maximum error is approximately \SI{0.06}{\degreeCelsius}, which is small and acceptable. 

Second, as mentioned in Section \ref{sec:Home energy management in PCM buildings}, the final value function \eqref{eqn23} consists of two parts: the electricity cost and the discomfort cost. For each, the MAE is calculated separately. The results show that the MAE of the electricity cost for $\lambda = 0.95$ is 0.74 cents, relative to an average value of 8.08 cents. For the discomfort cost, the MAE is about \SI{0.07}{\degreeCelsius} relative to an average of \SI{1.6}{\degreeCelsius}. These results indicate that the error of the final value function is acceptable.

Third, the approximation is also verified further by examining the calibration error between the optimal policy of the exact and approximated models over a 24-hour horizon. 
To calculate the calibration error for each starting point, the difference in the number of on-cycles in the equivalent optimal policy is divided by a total number of time steps (24), and the results are averaged over the 61 starting points. 
Comparing the two cases shows that the difference in the number of on-cycles in the optimal policies is zero in 60 out of the total 61 studied cases. 
In the last case, the difference is one on-cycle, giving a calibration error of \SI{0.07}{\%}, which is acceptable. 

Overall, these results demonstrate the efficacy of the state-space approximation, particularly in terms of model accuracy, which permits us to move on to \textsc{Alg}~1 and \textsc{Alg} 2.

\subsection{Evaluation of Algorithm~1 and Algorithm~2}
\label{subsec:Alg1vsAlg2}

\begin{table*}[t]
\scriptsize
	\caption{Evaluation of the proposed method.}
	\label{table1}
	\centering
\begin{tabular}{cllcccccccc}
\cline{2-11}
\multicolumn{1}{l|}{}                                                                                                    & \multicolumn{2}{c|}{Season}                                                                                                                              & \multicolumn{2}{c|}{Spring}                               & \multicolumn{2}{c|}{Summer}                               & \multicolumn{2}{c|}{Autumn}                               & \multicolumn{2}{c|}{Winter}                               \\ \hline
\multicolumn{1}{|c|}{\multirow{3}{*}{\begin{tabular}[c]{@{}c@{}}HVAC system with\\ deadband relay\end{tabular}}} & \multicolumn{2}{l|}{\begin{tabular}[c]{@{}l@{}}Average number of HVAC operating hours\end{tabular}}                                                       & \multicolumn{2}{c|}{26}                                   & \multicolumn{2}{c|}{12}                                   & \multicolumn{2}{c|}{14}                                   & \multicolumn{2}{c|}{93}                                   \\ \cline{2-11} 
\multicolumn{1}{|c|}{}                                                                                              & \multicolumn{1}{l|}{\multirow{2}{*}{\begin{tabular}[c]{@{}l@{}}Average cumulative  cost\end{tabular}}}  & \multicolumn{1}{l|}{Electricity cost (\$)} & \multicolumn{2}{c|}{14.62}                                & \multicolumn{2}{c|}{11.14}                                & \multicolumn{2}{c|}{7.69}                                 & \multicolumn{2}{c|}{57.24}                                \\ \cline{3-11} 
\multicolumn{1}{|c|}{}                                                                                              & \multicolumn{1}{l|}{}                                                                                              & \multicolumn{1}{l|}{Discomfort (\SI{}{\degreeCelsius})}   & \multicolumn{2}{c|}{325.24}                               & \multicolumn{2}{c|}{301.47}                               & \multicolumn{2}{c|}{251.74}                               & \multicolumn{2}{c|}{200.71}                                \\ \hline
\multicolumn{1}{l}{}                                                                                                &                                                                                                                    &                                     & \multicolumn{1}{l}{}        & \multicolumn{1}{l}{}        & \multicolumn{1}{l}{}        & \multicolumn{1}{l}{}        & \multicolumn{1}{l}{}        & \multicolumn{1}{l}{}        & \multicolumn{1}{l}{}        & \multicolumn{1}{l}{}        \\ \cline{4-11} 
\multicolumn{1}{l}{}                                                                                                &                                                                                                                    & \multicolumn{1}{l|}{}               & \multicolumn{4}{c|}{$\lambda=0.95$}                                                                                           & \multicolumn{4}{c|}{$\lambda=0.05$}                                                                                           \\ \cline{4-11} 
\multicolumn{1}{l}{}                                                                                                &                                                                                                                    & \multicolumn{1}{l|}{}               & \multicolumn{1}{l|}{Spring} & \multicolumn{1}{l|}{Summer} & \multicolumn{1}{l|}{Autumn} & \multicolumn{1}{l|}{Winter} & \multicolumn{1}{l|}{Spring} & \multicolumn{1}{l|}{Summer} & \multicolumn{1}{l|}{Autumn} & \multicolumn{1}{l|}{Winter} \\ \hline
\multicolumn{1}{|c|}{\multirow{3}{*}{ALG 1}}                                                                        & \multicolumn{2}{l|}{\begin{tabular}[c]{@{}l@{}}Average number of HVAC operating hours\end{tabular}}                                                       & \multicolumn{1}{c|}{23}     & \multicolumn{1}{c|}{19}     & \multicolumn{1}{c|}{11}     & \multicolumn{1}{c|}{74}     & \multicolumn{1}{c|}{34}     & \multicolumn{1}{c|}{28}     & \multicolumn{1}{c|}{38}     & \multicolumn{1}{c|}{71}     \\ \cline{2-11} 
\multicolumn{1}{|c|}{}                                                                                              & \multicolumn{1}{l|}{\multirow{2}{*}{\begin{tabular}[c]{@{}l@{}}Average cumulative  cost\end{tabular}}} & \multicolumn{1}{l|}{Electricity cost (\$)} & \multicolumn{1}{c|}{11.53}  & \multicolumn{1}{c|}{8.20}    & \multicolumn{1}{c|}{4.48}   & \multicolumn{1}{c|}{40.89}  & \multicolumn{1}{c|}{21.64}  & \multicolumn{1}{c|}{22.86}  & \multicolumn{1}{c|}{19.20}   & \multicolumn{1}{c|}{46.21}   \\ \cline{3-11} 
\multicolumn{1}{|c|}{}                                                                                              & \multicolumn{1}{l|}{}                                                                                              & \multicolumn{1}{l|}{Discomfort (\SI{}{\degreeCelsius})}   & \multicolumn{1}{c|}{202.14} & \multicolumn{1}{c|}{112.08} & \multicolumn{1}{c|}{258.95} & \multicolumn{1}{c|}{155.44} & \multicolumn{1}{c|}{156.36} & \multicolumn{1}{c|}{37.79}  & \multicolumn{1}{c|}{218.37} & \multicolumn{1}{c|}{110.41}  \\ \hline
\multicolumn{1}{l}{}                                                                                                &                                                                                                                    &                                     & \multicolumn{1}{l}{}        & \multicolumn{1}{l}{}        & \multicolumn{1}{l}{}        & \multicolumn{1}{l}{}        & \multicolumn{1}{l}{}        & \multicolumn{1}{l}{}        & \multicolumn{1}{l}{}        & \multicolumn{1}{l}{}        \\ \cline{4-11} 
\multicolumn{1}{l}{}                                                                                                &                                                                                                                    & \multicolumn{1}{l|}{}               & \multicolumn{4}{c|}{$\lambda=0.95$}                                                                                           & \multicolumn{4}{c|}{$\lambda=0.05$}                                                                                           \\ \cline{4-11} 
\multicolumn{1}{l}{}                                                                                                &                                                                                                                    & \multicolumn{1}{l|}{}               & \multicolumn{1}{l|}{Spring} & \multicolumn{1}{l|}{Summer} & \multicolumn{1}{l|}{Autumn} & \multicolumn{1}{l|}{Winter} & \multicolumn{1}{l|}{Spring} & \multicolumn{1}{l|}{Summer} & \multicolumn{1}{l|}{Autumn} & \multicolumn{1}{l|}{Winter} \\ \hline
\multicolumn{1}{|c|}{\multirow{4}{*}{ALG 2 against ALG 1}}                                                          & \multicolumn{2}{l|}{Max. MAE of indoor temp. of optimal policy (\SI{}{\degreeCelsius})}         &                                                     \multicolumn{1}{c|}{1.5}   & \multicolumn{1}{c|}{1.6}    & \multicolumn{1}{c|}{1.2}   & \multicolumn{1}{c|}{2.0}   & \multicolumn{1}{c|}{1.3}   & \multicolumn{1}{c|}{1.1}   & \multicolumn{1}{c|}{2.1}   & \multicolumn{1}{c|}{1.3}   \\ \cline{2-11} 
\multicolumn{1}{|c|}{}                                                                                              & \multicolumn{1}{l|}{\multirow{2}{*}{\begin{tabular}[c]{@{}l@{}}MAE of the cumulative cost\end{tabular}}}  & \multicolumn{1}{l|}{Elect. cost error (\%)}      & \multicolumn{1}{c|}{7.89}   & \multicolumn{1}{c|}{6.71}   & \multicolumn{1}{c|}{14.95}   & \multicolumn{1}{c|}{6.11}    & \multicolumn{1}{c|}{2.86}   & \multicolumn{1}{c|}{1.88}   & \multicolumn{1}{c|}{20.83}      & \multicolumn{1}{c|}{0.84}   \\ \cline{3-11} 
\multicolumn{1}{|c|}{}                                                                                              & \multicolumn{1}{l|}{}                                                                                              & \multicolumn{1}{l|}{Discomfort error (\%)}   & \multicolumn{1}{c|}{7.14}  & \multicolumn{1}{c|}{13.94}  & \multicolumn{1}{c|}{5.48}  & \multicolumn{1}{c|}{10.63}  & \multicolumn{1}{c|}{6.38}   & \multicolumn{1}{c|}{15.85}   & \multicolumn{1}{c|}{2.93}    & \multicolumn{1}{c|}{19.95}  \\ \cline{2-11} 
\multicolumn{1}{|c|}{}                                                                                              & \multicolumn{2}{l|}{\begin{tabular}[c]{@{}l@{}}Normalised calibration error (\%)\end{tabular}}                                       & \multicolumn{1}{c|}{1.39}    & \multicolumn{1}{c|}{0.49}   & \multicolumn{1}{c|}{0.66}    & \multicolumn{1}{c|}{0.95}      & \multicolumn{1}{c|}{0.61}    & \multicolumn{1}{c|}{0.54}    & \multicolumn{1}{c|}{3.59}    & \multicolumn{1}{c|}{0.96}    \\ \hline
\end{tabular}
\end{table*}

\begin{figure}[t]
	\centering
	\includegraphics[width=90mm,keepaspectratio]{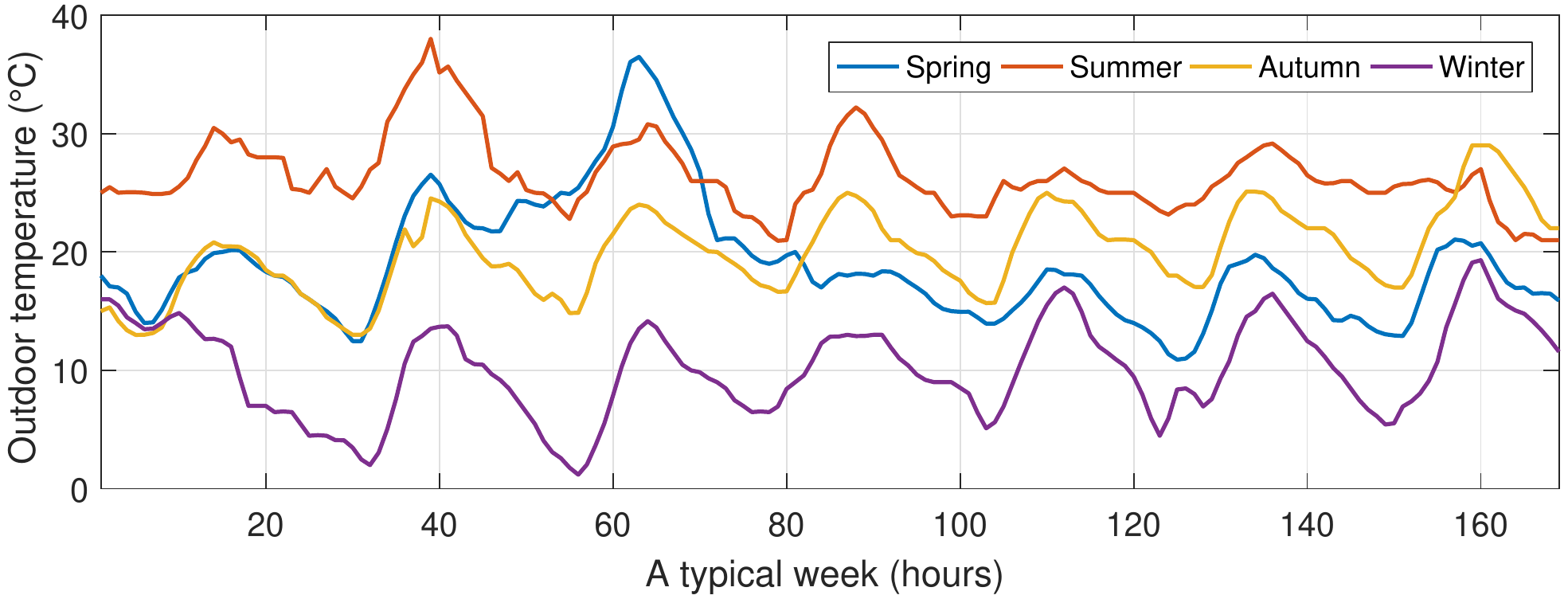}
	\caption{Outdoor temperature of four different seasonal weather conditions in Sydney: spring (16/10-22/10), summer (16/01-22/01), autumn (1/04-7/04) and winter (25/07-31/07).}
	\vspace{-0.2cm}
	\label{fig:season}
\end{figure}

We now evaluate the proposed methodology, \textsc{Alg}~2, 
and the approach using only the multiple timescales abstraction, \textsc{Alg}~1, in order to ascertain the benefits of the multi-timescale and macro-action abstractions. Moreover, to get a better sense of the benefits of using optimal HVAC scheduling in PCM-buildings, the algorithms are compared to a simple deadband relay for controlling the HVAC system.  Furthermore, to demonstrate the ability of the two algorithms to capture the customer preferences in terms of electricity cost and comfort, the algorithms are run for two different weighting factors: $\lambda=0.95$ (more weight on the electricity cost) and $\lambda=0.05$ (more weight on the thermal discomfort). The results for four typical weeks, one for each season, are summarized in Table~\ref{table1}. The temperature profiles of the four weeks are shown in Fig.~\ref{fig:season}.

\subsubsection{Deadband policy vs. \textsc{Alg}~1}
To begin, we consider a conventional HVAC system operating with a deadband control.
We simulate an identical home with an identical HVAC system. The only difference is that the HVAC system is controlled using a deadband controller as opposed to optimal scheduling used in \textsc{Alg}~1 and \textsc{Alg}~2.
The deadband range is set between \SI{20}{\degreeCelsius} and \SI{22}{\degreeCelsius} for heating, and between \SI{22}{\degreeCelsius} and \SI{26}{\degreeCelsius} for cooling.  The simulations are run for 61 initial points. We record the average number of HVAC operating hours and the average cumulative electricity cost, and compare the performance of the HVAC system with a deadband relay against \textsc{Alg}~1.

The results show considerable benefits from using optimization over simple deadband control.
Specifically, using \textsc{Alg}~1 for  a typical winter week with a weighting factor of $\lambda=0.05$ reduces the number of HVAC operating hours by \SI{23.7}{\%}.
Importantly, both cost function components see improvements: the electricity cost decreases by \SI{19.3}{\%}, while the number of discomfort hours also decreases by \SI{45}{\%}. 
In contrast, in summer, applying \textsc{Alg}~1 increases the number of HVAC operating hours by a factor of 2.3, which increases the electricity cost by a factor of two. The number of discomfort hours, on the other hand, reduces by \SI{87.5}{\%}, which is to be expected given that a much higher weight ($\lambda=0.05$) is given to the comfort of the occupants. Putting more weight on the electricity cost ($\lambda=0.95$) \textit{reduces} the electricity cost by \SI{26}{\%} at the expense of a slightly lower reduction in the thermal discomfort (now \SI{63}{\%}), which goes to show that the weighting factor $\lambda$ has to be carefully tuned for optimal performance with respect to customers' preferences.

The only season when $\lambda=0.95$ doesn't result in the reduction in both cost components is Autumn. However, the increase in discomfort is only \SI{3}{\%} while the reduction in electricity cost is significant (about \SI{40}{\%}), which confirms the superiority of \textsc{Alg}~1 over the deadband control.


\subsubsection{\textsc{Alg}~1 vs. \textsc{Alg}~2}Finally, we evaluate \textsc{Alg}~2 with respect to \textsc{Alg}~1, using as a comparison metrics: (i) maximum MAE of the indoor temperature; (ii) MAE of the cumulative cost expressed as a percentage deviation from \textsc{Alg}~1; and, (iii) the normalized calibration error. 
We compute the first measure only for the optimal policy rather than for the whole state-space, because using macro actions in \textsc{Alg}~2, results in a different size of the state-space compared to \textsc{Alg}~1. 

The results for the first metric show that the maximum MAE of the indoor temperature over four typical weeks for $\lambda=0.95$ and  $\lambda=0.05$, are \SI{2}{\degreeCelsius} and  \SI{2.1}{\degreeCelsius}, respectively, which is about \SI{0.01}{\degreeCelsius} on average per hour. 
For the second metric, using macro-action abstraction in \textsc{Alg~2} reduces the solution quality compared to \textsc{Alg~1} by up to \SI{20}{\%} at most (electricity cost in Autumn and discomfort in Winter, both for $\lambda=0.05$), but is otherwise mostly below \SI{10}{\%}.
The third metric, the maximum calibration error, over four case studies is \SI{1.39}{\%} and \SI{3.59}{\%} for $\lambda=0.95$ and $\lambda=0.05$, respectively. This implies that the optimal policies from \textsc{Alg}~2 and \textsc{Alg}~1 have a very similar number of on-cycles.

Overall, the results of these three metrics show that the performance of \textsc{Alg}~2 is comparable compared to \textsc{Alg}~1. 
Note that there is a trade-off between the superior runtime of \textsc{Alg}~2 and a better accuracy of \textsc{Alg}~1 in tracking the optimal policy. 
These errors may be mitigated by adding extra macro actions, but this will in turn reduce the computational efficiency of \textsc{Alg}~2. 

\begin{figure}[t]
	\centering
	\includegraphics[width=90mm,keepaspectratio]{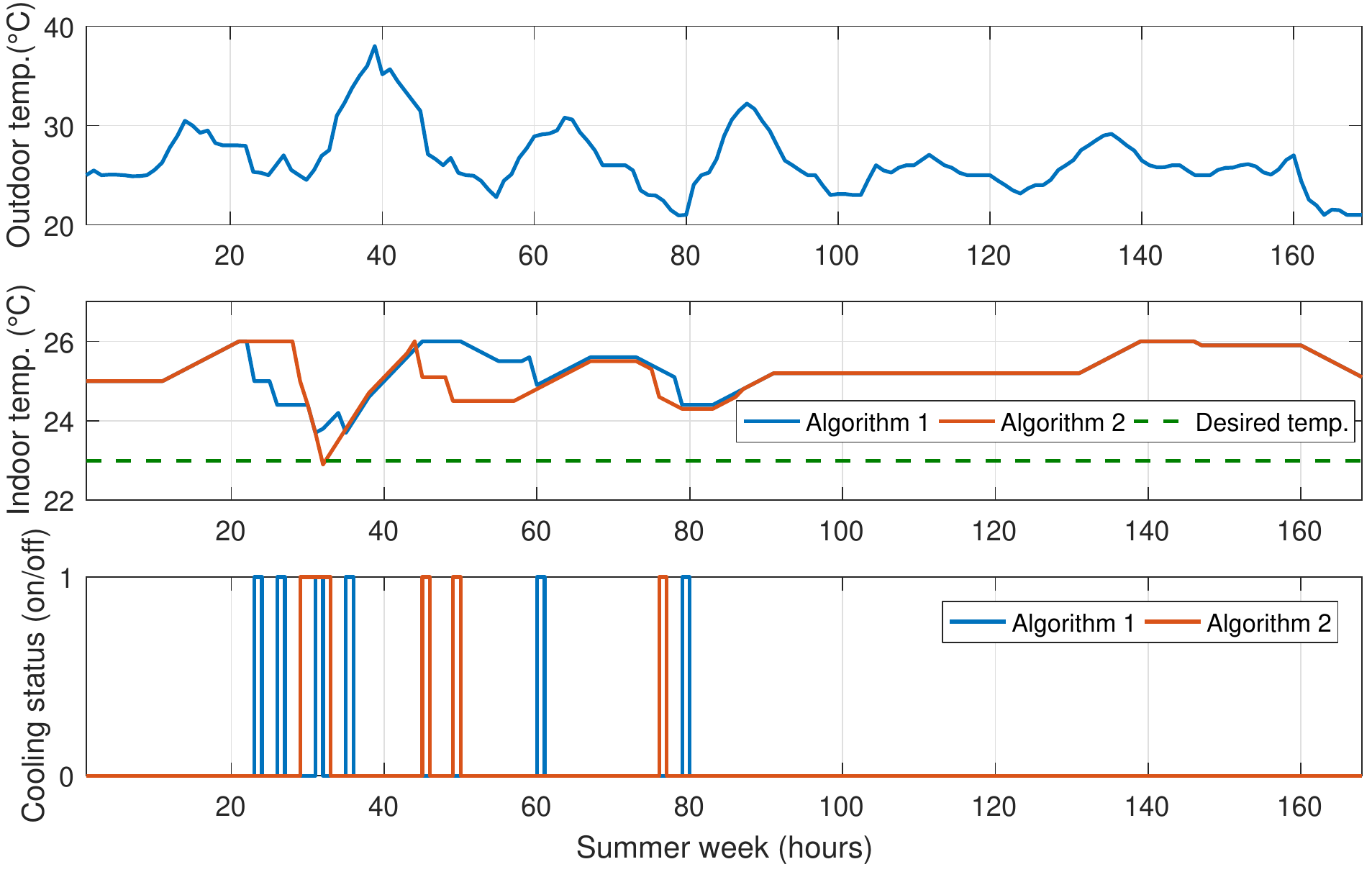}
	\caption{From top to bottom: outdoor temperature over a typical summer week in Sydney; indoor temperature corresponds to an optimal scheduling of HVAC system that results from applying the \textsc{Alg}~1 versus the \textsc{Alg}~2 and finally optimal operation of HVAC system that results from applying the \textsc{Alg}~1 versus the \textsc{Alg}~2. The dash line represents customer's desired temperature in a summer week.}
	\vspace{-0.2cm}
	\label{fig:sim}
\end{figure}

To visually compare the performance of the two algorithms, Fig.~\ref{fig:sim} shows the indoor temperature and the HVAC schedule for the two algorithms for a typical summer week. 
Observe that the both temperature profiles are within the desired comfort range, with only a slight difference in the second day when the the on/off schedules of the HVAC system don't match.
It is worth mentioning that the indoor temperature for some hours such as 0-10 and 90-130, is constant, and this where phase change occurs. During this phase change, the PCM absorbs the heat from the building's interior and keeps the indoor temperature of the building within the desired range.

We have also compared the runtime of both algorithms to be able to quantify the computational cost saving resulting from the use of  macro-action and multi-timescale abstractions. We can observe a speedup of up to 12,900 times compared to the direct application of DP. The use of macro actions in \textsc{Alg}~2 results in a speedup of up to 2.4 times compared to \textsc{Alg~1}.

\section{Conclusion} 
\label{sec:Conclusion}

In this work, we have addressed the computational challenge of solving an optimization problem in buildings with PCM.
We have developed a computationally efficient macro-action multi-timescale algorithm to deal with the computational burden of a large non-linear, non-convex HVAC scheduling problem. 
We demonstrated the efficacy of the proposed approach on a typical PCM-building over four typical weeks that are representative of four seasons in Sydney. 
The results demonstrate the superior computational performance of the proposed scheduling algorithm compared to a direct application of DP while maintaining an acceptable solution quality. The results also show that the weighting of the electricity and the discomfort costs in the objective function have to be selected carefully to ensure an optimal trade-off between electricity expenditure and thermal discomfort.

In future work, we will consider stochastic exogenous inputs, such as weather conditions and occupants' behavior, which in this paper were considered deterministic.

\vspace{-0.2cm}
\bibliographystyle{IEEEtran}
\bibliography{JZR}

\newpage

\section*{Appendix A}
\label{sec:Appendix A}
Here we briefly explain the proof of the multi-timescale approach presented Section \ref{sec:Methodology}-B. The proof is based on a generalized Bellman equation \cite{sutton1995td}
\begin{align}\label{eqn26} 
V= C+ \mathbb{P}^{T}V.
\end{align}
The model is valid if it satisfies \eqref{eqn26} for any $\mathbb{P}$ and $C$, with $\lim_{i \to \infty} \mathbb{P}^{i}=0$, where $i$ is the number of the MDPs in the model. For any valid model, we can update the value function through lookahead or backup operation as follows:
\begin{align}\label{eqn27} 
V_{k+1}= C+ \mathbb{P}^{T}V_{k}.
\end{align}
 As long as the model is valid it converges to the same value function regardless of the number of steps: 
\begin{align}\label{eqn28} 
V_{\infty}= \sum^{\infty}_{i=0} \mathbb{P}^{T^{i}}C=V
\end{align}
To prove that the solution of a multi-timescale MDP is the same as the solution of an one-step MDP, we need to prove that an $i$-step model formulation satisfies the generalized Bellman equation \eqref{eqn26}.  
\par\vspace{3ex}
\noindent\textbf{Theorem \thinspace\Rmnum{6.1}}: A multi timescale or an $n$-step model that has a general form \eqref{eqn29} satisfies the generalized Bellman equation \eqref{eqn26}.
\begin{subequations} \label{eqn29}
	\begin{align}
	\begin{split} \label{eqn29c}
	{C}^{(n)}= \sum^{n-1}_{i=0} \left(\mathbb{P}^{T}\right)^{i}C,\\
	\end{split}\\
	\begin{split} \label{eqn29d}
	{C}^{(n)^{T}}{s}_{k}= \mathbb{E}(c^{(n)}_{k}|{s}_{k}),\\
	\end{split}
	\end{align}
	\label{straincomponent}
\end{subequations}
\hspace{-3.5pt}where ${c}^{(n)}_{k}= \sum^{n}_{i=1}{c}_{k+i}$ is the n-step truncated return starting from state ${s}_{k}$. 
\vspace{6pt}
 \par \noindent\textbf{Proof}: We combine $\mathbb{P}$ and $C$ and the initial value ${s}_{0}$ into a matrix $M$:
\[ M=
\left(\begin{array}{@{}c|cccc@{}}
{s}_{0} & C^{T} \\\hline
0 & \mathbb{P}  \\
\end{array}\right)
\]
If the vector $V$ is also augmented by adding an initial component whose value is always $1$, then the generalized Bellman equation \eqref{eqn26}, can be written as
\begin{align}\label{eqn30} 
V=M^{T}V.
\end{align}
Same as before, we consider model $M$ to be valid if and only if it satisfies \eqref{eqn30}. For any valid model $M_{i}$, the composed model $\prod\limits_{i=1}^{n}{M}_{i}$ is also valid because
\begin{align}\label{eqn31} 
\prod_{i=1}^{n}\left({M}_{i}\right)^{T}V=\prod_{i=1}^{n}{M}_{i}^{T}V=V
\end{align}
Note that $M$ has been constructed such that it is valid only if the corresponding $\mathbb{P}$ and $C$ are valid. Therefore, \eqref{eqn31} proves the validity of the $n$-step model \eqref{eqn29}. \hspace{89pt} $\blacksquare$



\end{document}